\documentclass[runningheads,a4paper]{llncs}

\usepackage{amssymb}
\usepackage{graphicx}

\usepackage{url}
\urldef{\mailsa}\path|avterekhov@gmail.com, montone.guglielmo@gmail.com,|
\urldef{\mailsb}\path|kevin.oregan@parisdescartes.fr|    
\newcommand{\keywords}[1]{\par\addvspace\baselineskip
\noindent\keywordname\enspace\ignorespaces#1}

\begin{document}

\mainmatter  % start of an individual contribution

\title{Knowledge transfer in deep block-modular neural networks}

\titlerunning{Knowledge transfer in deep block-modular neural networks}

% \author{Alfred Hofmann%
% \thanks{Please note that the LNCS Editorial assumes that all authors have used
% the western naming convention, with given names preceding surnames. This determines
% the structure of the names in the running heads and the author index.}%
% \and Ursula Barth\and Ingrid Haas\and Frank Holzwarth\and\\
% Anna Kramer\and Leonie Kunz\and Christine Rei\ss\and\\
% Nicole Sator\and Erika Siebert-Cole\and Peter Stra\ss er}
\author{Alexander V. Terekhov \and Guglielmo Montone \and J. Kevin O'Regan}
\authorrunning{A.V.~Terekhov, G.~Montone, J.K.~O'Regan}

\institute{Laboratoire Psychologie de la Perception\\Universit\'{e} Paris Descartes\\75006 Paris, France\\
\mailsa\\
\mailsb\\
\url{http://lpp.psycho.univ-paris5.fr/feel}}


\toctitle{Title}
\tocauthor{A.V. Terekhov, G. Montone and J.K. O'Regan}
\maketitle

\begin{abstract}

Although deep neural networks (DNNs) have demonstrated impressive results during the last decade, they remain highly specialized tools, which are trained -- often from scratch -- to solve each particular task. The human brain, in contrast, significantly re-uses existing capacities when learning to solve new tasks. In the current study we explore a block-modular architecture for DNNs, which allows parts of the existing network to be re-used to solve a new task without a decrease in performance when solving the original task. We show that networks with such architectures can outperform networks trained from scratch, or perform comparably, while having to learn nearly 10 times fewer weights than the networks trained from scratch.
\end{abstract}

\section{Introduction}

Deep Neural Networks (DNN) have demonstrated impressive results in the last 10-15 years. They have established new benchmarks in such tasks as the classification of hand-written digits  \cite{hinton2006fast}, object recognition \cite{lee2009convolutional}, speech recognition \cite{graves2013speech}, machine translation \cite{sutskever2014sequence} and many others. These successes can mainly be attributed to the use of pre-training \cite{erhan2009difficulty,salakhutdinov2012efficient},  sharing \cite{lecun1995convolutional}, and various forms of regularization \cite{girosi1995regularization,glorot2011deep}, as well as to  increases in computational capacity and data availability \cite{schmidhuber2015deep}.

In spite of these impressive results, DNNs are still unable to match humans in the diversity of tasks we can solve and the ease  which we learn. For example, we are able to progressively accumulate and abstract knowledge from previous experiences and re-use this knowledge to perform new tasks. On the contrary, in DNNs learning a new task tends to erase the knowledge about the previous tasks --- a phenomenon known as \textit{catastrophic forgetting} \cite{mccloskey1989catastrophic}.

This specialization contrasts with the operating principles of the human nervous system, as it is well known that the brain re-uses existing structures when learning a new task \cite{anderson2010neural,dehaene2007cultural}. Ideally, we would like neural networks to possess similar capacities and operating principles. Imagine a neural network NN$_1$ has been trained on task T$_1$, and we would like to have a new network NN$_2$ which solves task T$_2$, where T$_1$ and T$_2$ share common features. If we simply take the network NN$_1$ and train it on T$_2$, it will most probably perform more poorly when solving T$_1$ after retraining. Of course, we can make an exact copy of NN$_1$ and then train it on T$_2$, but this is computationally cumbersome. Moreover, when using such a paradigm, we will simply obtain a collection of networks which are all very specialized and all work independently from one another. Rather, we would like to have a system that re-uses features learned in task T$_1$ to solve task T$_2$, and eventually, after having learned a number of tasks,  would re-use relevant features from all (or more likely some) previous tasks to solve a new task.

The idea of exploiting features learned in a previous task when solving a new task is not original in the field of artificial neural networks. Simultaneous learning of multiple related tasks can lead to improves the performance in DNNs \cite{collobert2008unified} especially if the training data is limited in one of the tasks \cite{liurepresentation}. The studies with sequential learning of multiple tasks are rare.
In work by Gutstein and colleagues \cite{gutstein2008knowledge} a multi-layer convolutional neural net was trained to recognize a set of digits. It was then shown that the same network, when trained on a second set of digits, achieved better performance when only the upper layers of the network -- and not the entire network -- were re-trained on the second task. Clearly the main problem with this kind of architecture is that learning a new task necessarily undermines those previously learned~\cite{goodfellow2013Empirical}. The problem of solving several tasks has been addressed in modular neural networks. These networks are able to detect different training patterns, corresponding to different tasks, and allocate different sub-networks to learn them \cite{jordan1991competitive,lu1999task}.

In the current paper we explore an alternative approach to training DNNs, partly inspired by modular NNs. This approach will allow the network to learn a new task by exploiting previously learned features. At the same time, the learning procedure is such that training on a new task will not affect the performance of the network on the previously learned tasks. Our procedure consists of training an initial network on a task, and then, instead of copying the network and training it on a new task, adding blocks of neurons to the original network and learning the connections between the neurons in the original network and the neurons in the introduced blocks. We repeat this procedure on multiple tasks, showing that the final architecture is able to learn a new task by adding a rather small number of blocks of neurons and connection weights to the original network, when compared with a number of neurons and connection weights in a network which must learn the new task from scratch.
    
The paper is organized as follows: in the next section we describe the structure of the neural networks used and the techniques used to add blocks of neurons to such networks. In the Methods section we present the tasks on which we train different networks. We then provide details on the networks and on the learning algorithm. Finally, in the Results section we report the performance of the networks which learned from scratch and those which learned by adding blocks of neurons to existing networks.

\subsection{Block-modular network architecture}

Consider a neural network with an input layer, multiple hidden layers, and an output layer such as the one presented in Figure 1a. Such a network, after being trained on a certain task, T$_1$, has definite values of weights and biases. We create a new network by adding neurons to each of the layers, including the output layer, as shown in Figure 1b. We refer to the added neurons as block neurons. 

The first layer block neurons receive projections from the input only, and in this aspect are qualitatively similar to the original first-layer neurons. The second-layer hidden block neurons receive inputs from both the first layer original neurons and the first layer block neurons. This pattern is repeated for all hidden layers. The output layer block neurons receive inputs from the last hidden layer of both original and block neurons.

In the current study we explore the applicability of such an architecture to classification tasks. We use softmax units in the output layer, which we refer to as the classification layer. Note that the softmax is computed independently for the original and block classification layer neurons. We train the original network on a task T$_1$ and then train the weights of the block neurons on a task T$_2$. The resulting network is then able to perform both tasks T$_1$ and T$_2$. Note that such a network has two classifiers: one for T$_1$ (original classification neurons) and one for T$_2$ (block classification neurons).

We also consider variations of this type of block network. Most frequently, we add blocks to all layers, except for the first hidden layer, as illustrated in Figure 1c. In this situation the neural network does not receive raw input information; its only inputs are the outputs of the first hidden layer of the original network. As in the previous case (Figure 1b), only the weights to the block neurons are learned.

Additionally, we make use of several original networks, trained for different tasks. An example of adding blocks to a pair of original networks is presented in Figure~1d. Both original networks remain unaffected by the introduction of new neurons. The block neurons receive inputs from both of the original networks, and only connection weights to the block neurons are changed when learning a new task.

\begin{figure}
\centering
\includegraphics[width=\linewidth]{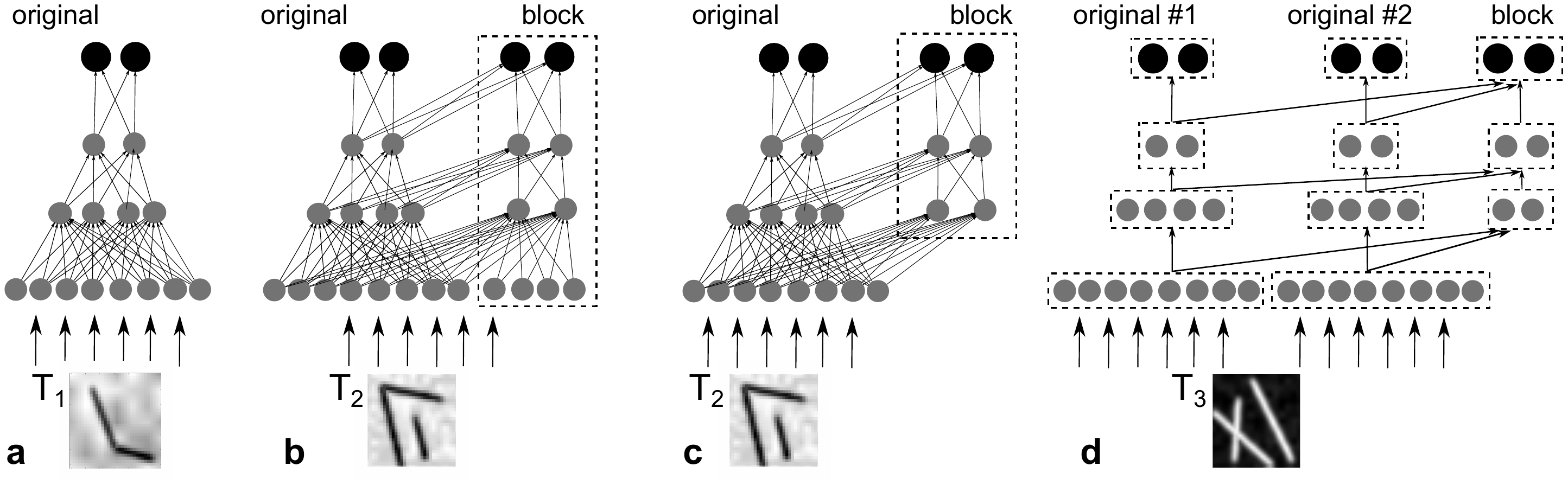}
\caption{(a) The architecture of the original network. We use a feedforward network with three hidden layers. (b) An additional block of neurons is added to each layer of the network. This is represented by the neurons in a dashed box. Each layer of the additional block is fully connected with the layer directly above and/or below within the block itself and with the layer below in the original network. (c) One of the architectures most used in the present work. The architecture is based on an original network trained on a first task T$_1$. A block of two hidden layers was added to the original network. The neurons were added at the second and third hidden layers only. The resulting architecture was trained on a second task T$_2$. (d) Adding blocks to two original networks. The dashed boxes indicate the layers of the two original networks and the blocks added. An arrow connecting two boxes indicates that all the neurons in the first box are connected to all the neurons in the second box.}
\label{fig:example}
\end{figure}

\section{Methods}

\subsection{Tasks}
The driving force behind our suggested architectures (e.g. Figure 1b-d) is the idea that the added neurons (being trained on task T$_2$) will re-use the capacities of the original neurons (trained on task T$_1$) whenever these capacities are relevant to the new task (T$_2$). Of course, such re-use will be most efficient when the tasks T$_1$ and T$_2$ have something in common. 

In order to explore the possibility of the re-use of network capacities we designed several tasks which, to a human observer, involve the notions of line and angle. Specifically, we designed six tasks, as illustrated in Figure 2.

In each task the stimuli were gray scale images, 32 x 32 pixels in size. Each image contained two to four line segments, each at least 13 pixels long (30\% of the image diagonal). The distance between the end points of each line segment and every other line segment was at least 4 pixels (10\% of the image diagonal). In order to obtain anti-aliased images, the lines were first generated on a grid three times larger (96 x 96). The images were then filtered with a Gaussian filter with sigma equal to 3 pixels, and downsampled to the final dimensions of 32~x~32 pixels. 

We used the following conditions (see illustrations in Figure 2).

\textit{ang\_crs:} requires classifying the images into those containing an angle (between 20$^\circ$ and 160$^\circ$) and a pair of crossing line segments (the crossing point must lay between 20\% and 80\% along each segment's length).

\textit{ang\_crs\_line:} the same as \textit{ang\_crs}, but has an addition line segment crossing neither of the other line segments.

\textit{ang\_tri\_ln:} distinguishes between images containing an angle (between 20$^\circ$ and 160$^\circ$) and a triangle (with each angle between 20$^\circ$ and 160$^\circ$ ); each image also contains a line segment crossing neither angle nor triangle.

\textit{blnt\_shrp:} requires classifying the images into those having blunt (between 100$^\circ$ and 160$^\circ$) and those having sharp (between 20$^\circ$ and 80$^\circ$) angles in them.

\textit{blnt\_shrp\_ln:} the same as \textit{blnt\_shrp}, but has an additional line segment, crossing neither of the line segments forming the angle.

\textit{crs\_ncrs:} distinguishes between a pair of crossing and a pair of non-crossing lines (the crossing point must lay between 20\% and 80\% of each segment length).

\begin{figure}[t]
\centering
\includegraphics[width=\linewidth]{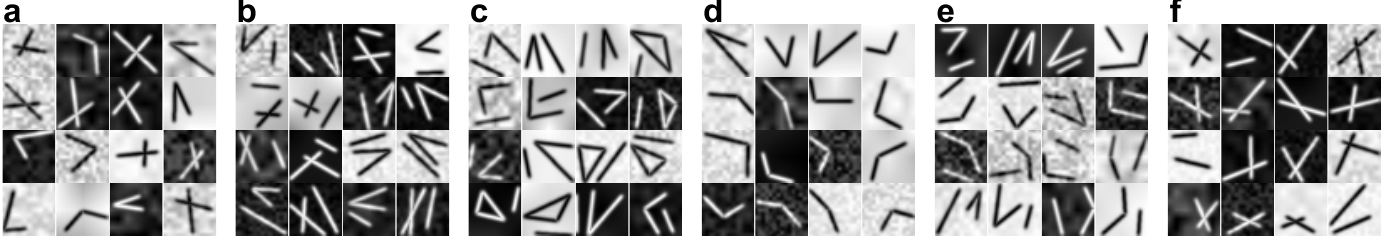}
\caption{Examples of stimuli: (a) \textit{ang\_crs} -- line segments forming an angle vs. two crossing line segments; (b) \textit{ang\_crs\_line} -- same, with an additional non-crossing line segment; (c) \textit{ang\_tri\_ln} -- angle vs. triangle; (d) \textit{blnt\_shrp} -- blunt angle vs. sharp angle; (e) \textit{blnt\_shrp\_ln} -- same, with a non-crossing line segment; (f) \textit{crs\_ncrs} -- two crossing line segments vs. two non-crossing line segments.
}
\label{fig:example}
\end{figure}

Each stimulus was generated by randomly selecting an appropriate number of points and verifying that all conditions were satisfied. Each image was then combined with a random background, $I_{background}$. Four different types of random background were generated with four patterns changing with different velocities. In particular a grid of step size $s$ with $s \in \left\lbrace 3,7,11,15 \right\rbrace$ was superimposed onto a 32~x~32 image. The values of the pixels corresponding to the grid nodes were randomly drawn from a uniform distribution between $0.1$ and $0.9$. The values of the remaining image pixels were obtained by linearly interpolating the randomly drawn ones. We used positive and negative stimuli. The positive stimuli were defined by the formula:
$$
I_{positive}(x, y) = \varepsilon I_{background} + (1 - \varepsilon)I_{stimulus},
$$
where $\varepsilon$ was randomly selected between $0.1$ and $0.4$ for each stimulus.

The negative stimuli were defined as $I_{negative} = 1 - I_{positive}$

For our experiments we generated 700,000 stimuli for each condition.

\subsection{Neural Network details}

\subsubsection{Original neural networks}

The original neural networks had three hidden layers and one classification layer with two softmax neurons:
$$
z_i = \frac{e^{x_i}}{e^{x_1} + e^{x_2}},
$$
where $z_i$ is the output of the $i$-th classification neuron ($i = 1,2$), and $x_i$ is the activation of the corresponding neuron.

Each of the hidden neurons had a rectified linear activation function:
$$
y = x_+ = \left\{
\begin{array}{lr}
x,& x \ge 0,\\
0,& x < 0.
\end{array}
\right.
$$
where $y$ is the output of this neuron and $x$ is its activation.

The activation of the $i$-th neuron at the $k$-th level (with $m^{(k)}$ neurons) was computed as a weighted sum of the outputs of neurons from the previous layer:
$$
x^{(k)}_i = \sum_{j=1}^{m^{(k)}} w^{(k)}_{ji} y_j^{(k-1)} + b_i^{(k)}
$$
Here $b_i^{(k)}$ is the bias.

For the first hidden layer, the activation was computed as a weighted sum of the inputs.

For every task, the original neural network had three hidden layers with 200, 100 and 50 neurons, respectively, and one classification layer with 2 neurons. The network received a vector of image pixels, which was of dimension 1024.

\subsubsection{Block neural networks} We used several block neural network configurations. The block neural network always had two softmax output neurons. The structure of the hidden layers is described by a triplet of numbers. For example, the triplet 100-50-25 represents 100 neurons added to the first layer, 50 to the second, and 25 to the third (see Figure 1b). In many cases the block network only received inputs from the first layer of the original network (see Figure 1c), rather than from the stimuli directly.  In these cases the first number in the triplet is zero, e.g. 0-50-50.

\subsubsection{Cost function}

The network was trained to minimize a cost function $J$ which combined three terms: the quality of the prediction $J_1$, the sparsity of the neurons' activation $J_2$, and the values of the weights $J_3$:
$$
J = J_1 + \beta J_2 + \lambda J_3
$$

The quality of the prediction was measured using the negative log-likelihood of the prediction given the data:
$$
J_1 = -\frac{1}{N}\sum_{n=1}^{N}\log z_{i(n)}(x_n^{(0)})
$$
where $x_n^{(0)}$ is the $n$-th training example, $z_{i(n)}$ is the output of the classifier corresponding to the correct answer.

The sparsity term of the cost function requires that each neuron in the hidden layer be active for $\rho N$ samples from the training data and silent otherwise. This measure was evaluated using the KL divergence~\cite{bradley2008differential}:
$$
J_2 = \sum_{k=1}^{M-1} \sum_{j=1}^{m_k} \rho\log\frac{\rho}{\rho^{(k)}_j} + (1-\rho)\log\frac{1-\rho}{1-\rho^{(k)}_j},
$$
where $\rho^{(k)}_j$ is the average activation of the $j$-th neuron in the $k$-th layer, $M$ is the number of layers (we do not apply this regularization to the classification layer), and $m^{(k)}$ is the number of neurons in the $k$-th layer.

The third term in the cost function is simply
$$
J_3 = \sum_{k=1}^{M-1} \sum_{j=1}^{m_k} \sum_{i=1}^{m_{k-1}} \left(w^{(k)}_{ij}\right)^2.
$$

The target sparseness $\rho$ was set to 0.05, and the coefficient $\beta$ was equal to 0.01. The weight-limiting coefficient $\lambda$ was set to 0.0001.

\subsubsection{Training}
The weights of the $k$-th hidden level were initialized with random uniformly distributed values in the range $\pm\sqrt{6/(n^{(k-1)}+n^{(k)})}$, where $n^{(k)}$ is the number of neurons in the $k$-th layer and $n^{(0)}$ is equal to the number of inputs. This initialization has been recommended for networks with $\tanh$ activation function~\cite{glorot2010under}; in the current study we used rectified linear units, but we kept the initalization range the same for compatibility with our pilot studies.

The total dataset was split into training (680,000 samples), validation (10,000 samples), and test (10,000 samples) datasets. The neural network was trained on the entire training dataset using mini-batch gradient descent learning with a batch size of 20. The initial update rate for the gradient descent was set to 0.01 and it decreased by a factor of 0.985 after every epoch. We used early stopping of the training process if the error on the validation dataset did not decrease after 5 epochs. The test score corresponding to the minimal validation error is presented as the performance of the network. Every condition was repeated 5~times, and the performance of the network was evaluated by the median value of the error.

\subsubsection{Implementation}
All code was written in python using Theano~\cite{bastien2012Theano,bergstra2010scipy}. Source files are available online: \url{https://github.com/feel-project/abstraction}

\section{Results}

We first present the results for the original networks, which were trained from scratch on each task. Then we present the results for the networks produced by adding blocks to the original networks and training the added weights on new tasks. We compare the performance of such networks to the original ones. Given the large number of permutations of weights and tasks, all possible combinations were not studied. Instead we tried to examine a sample set that gives an understanding of the performance of block-modular networks. 
\subsubsection{Original neural networks}

The performance of the original neural networks is presented in Table~1. These results show that adding a line to the image (e.g. Figure 2ad vs. 2be) made the task significantly more complicated. The first layer weights learned by the network are shown in Figure 3. Interestingly, visually, the weights are rather similar for \textit{blnt\_shrp} and \textit{blnt\_shrp\_ln}, as well as for \textit{ang\_crs} and \textit{ang\_crs\_ln}.

\begin{figure}
\centering
\includegraphics[width=\linewidth]{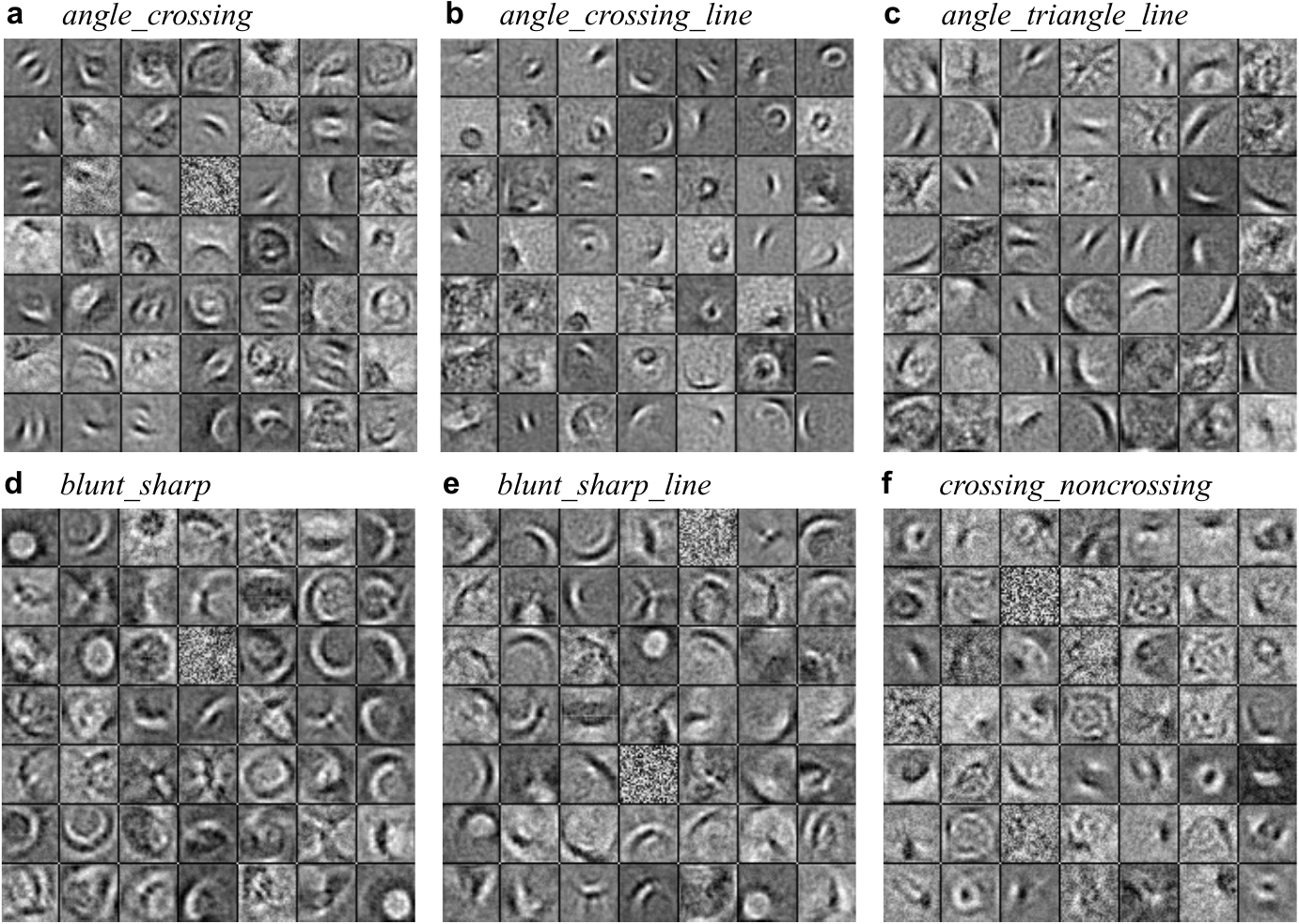}
\caption{Examples of weights. Weights (normalized) correspond to randomly selected first layer neurons of original networks trained to perform the corresponding tasks.}
\label{fig:example}
\end{figure}

\begin{table}
\label{table1}
\centering
\caption{The performance of the original 200-100-50 networks on different tasks.}
\begin{tabular}{l|c}
condition & performance\\
\hline
\textit{ang\_crs} & 3.1 (2.4--3.6)\\
\textit{ang\_crs\_ln} & 7.8 (6.7--9.0)\\
\textit{ang\_tri\_ln} & 3.6 (3.0--4.1)\\
\textit{blnt\_shrp} & 1.2 (0.8--1.4)\\
\textit{blnt\_shrp\_ln} & 4.1 (3.6--5.2)\\
\textit{crs\_ncrs} & 1.6 (1.3--2.6)\\
\end{tabular}

\medskip

The numbers correspond to median (min--max) percentage of misclassified examples.
\end{table}

\subsubsection{Adding 0-50-50 and 0-100-50 blocks to original networks}

We tested whether the networks performing \textit{ang\_crs} and \textit{blnt\_shrp} tasks could be re-used to perform \textit{ang\_crs\_ln} and \textit{blnt\_shrp\_ln} tasks by adding blocks to the original networks. In spite of the apparent similarity of the first-layer weights for both tasks (see Figure 3), adding block networks resulted in rather poor performance on tasks with lines (compare results in Table~1 and 2). The results improved slightly when more neurons were added at the second hidden layer. 

Re-using \textit{ang\_crs\_ln} for \textit{ang\_crs} and \textit{blnt\_shrp\_ln} for \textit{blnt\_shrp}, yielded comparable performance on \textit{ang\_crs} and improved performance on \textit{blnt\_shrp} compared to that of the corresponding original networks. This, however, could be partially attributed to the fact that the tasks are similar and, as such, re-using the original network on a new task in some sense increases the available training data.

We also attempted to re-use the network trained on one task to perform tasks that were more substantially different, such as \textit{ang\_crs\_ln} for \textit{ang\_tri\_ln}. In spite of certain similarities in the weights in both tasks (Figure 3bc), the performance was rather poor compared to that of the corresponding original network.

\begin{table}
\label{table2}
\centering
\caption{Adding blocks to original networks.}
\begin{tabular}{l|c|c|c|c}
condition & 0-50-50 & 0-100-50 & 50-50-50 & 100-50-50 \\\hline
\textit{ang\_crs\_ln} (\textit{ang\_crs}) & 9.0 (8.5--9.4) & 8.2 (7.8--8.5) & \textbf{7.7 (7.4--8.7)} & \textbf{7.5 (7.0--9.0)} \\
\textit{blnt\_shrp\_ln} (\textit{blnt\_shrp}) & 6.6 (6.0--7.1) & 5.9 (5.2-6.2) & 5.5 (5.2--5.9) & 5.0 (4.7--5.3) \\
\textit{ang\_crs} (\textit{ang\_crs\_ln}) & \textbf{2.7 (2.4--3.3)} & \textbf{2.7 (2.4--3.2)} & \textbf{2.9 (2.5--3.3)} & \textbf{2.8 (2.6--3.5)} \\
\textit{blnt\_shrp} (\textit{blnt\_shrp\_ln}) & \textbf{0.8 (0.7--1.0)} & \textbf{0.7 (0.6--0.8)} & \textbf{0.8 (0.7--0.9)} & \textbf{0.7 (0.7--0.9)} \\
\textit{ang\_tri\_ln} (\textit{ang\_crs\_ln}) & 6.7 (6.3--7.6)& 5.6 (5.0--6.6) & 4.8 (4.2--5.5) & 4.3 (3.8--4.9)\\
\textit{blnt\_shrp\_ln} (\textit{ang\_crs\_ln}) & 7.8 (7.4--8.2) & 6.6 (6.2--7.1) & 5.0 (4.6--5.1) & 4.3 (4.3--4.6)\\
\textit{ang\_crs\_ln} (\textit{ang\_tri\_ln}) & 12.4 (11.8--13.4) & 10.4 (10.1--11.2) & 8.9 (8.7--10.2) & \textbf{7.8 (7.4--8.6)}\\
\textit{ang\_crs\_ln} (\textit{blnt\_shrp\_ln}) & 13.0 (12.8--13.4) & 11.7 (10.4--12.8) & 10.0 (9.5--10.4) & 8.8 (8.5--9.6) \\
\end{tabular}

\medskip

The name of the task on which the block neurons were trained is followed by the name of the task on which the original networks were trained (in brackets). The cases when the block networks outperform the original ones are marked with bold font.
\end{table}

\subsubsection{Adding 50-50-50 and 100-50-50 blocks to original networks}

One potential reason for the poor performance of the block networks discussed above is that they did not receive stimuli as inputs, but only the outputs of the first layer of the original neural networks. To test this hypothesis we added blocks which had 50 or 100 neurons in the first layer in addition to neurons in the other layers. These results are presented in Table 2. This modification yielded substantial improvement, with some of the conditions performing better than the original networks.

\subsubsection{Adding blocks to pairs of networks}
The previous observation suggested that having a richer first hidden layer may improve the performance of block networks. We further explored this by creating block networks based on pairs of original networks (see Figure~1d). The results are presented in Table~3. Clearly, block networks with an empty first layer performed comparably to or better than the original networks. It must be noted that the block networks learned an order of magnitude fewer parameters. Each original network had approximately 10$^5$ parameters, while each block network had about 2$\cdot$10$^4$ which were not shared with original networks (for 0-50-50 blocks). Adding neurons to the first hidden layer provided additional improvement, as shown in the last column of Table 3.

\begin{table}
\label{table3}
\centering
\caption{Adding blocks to pairs of original networks.}
\begin{tabular}{l|c|c|c}
condition & 0-50-50 & 0-100-50 & 50-50-50\\\hline
\textit{ang\_crs\_ln} (\textit{ang\_tri\_ln}$+$\textit{crs\_ncrs}) & 8.8 (8.5--9.0) & 8.3 (7.3--8.4) & \textbf{7.7 (7.6--7.8)}\\
\textit{ang\_crs\_ln} (\textit{ang\_tri\_ln}$+$\textit{blnt\_shrp\_ln}) & 8.3 (7.9--9.2) & \textbf{7.5 (7.3--8.4)} & 7.9 (7.7--8.5)\\
\textit{blnt\_shrp} (\textit{ang\_tri\_ln}$+$\textit{crs\_ncrs}) & \textbf{1.0 (0.9--1.2)} & \textbf{0.9 (0.8--1.0)} & \textbf{0.9 (0.8--1.0)}\\
\textit{blnt\_shrp} (\textit{ang\_tri\_ln}$+$\textit{ang\_crs\_ln}) & \textbf{0.8 (0.7--0.9)} & \textbf{0.7 (0.7--0.8)} & \textbf{0.8 (0.6--0.9)} \\
\textit{blnt\_shrp} (\textit{ang\_tri\_ln}$+$\textit{blnt\_shrp\_ln}) & \textbf{0.6 (0.6--0.7)} & \textbf{0.6 (0.6--0.8)} & \textbf{0.6 (0.5--0.7)} \\
\textit{blnt\_shrp\_ln} (\textit{ang\_tri\_ln}$+$\textit{ang\_crs\_ln}) & \textbf{4.1 (3.9--4.4)} & \textbf{3.8 (3.2--4.1)} & \textbf{3.6 (3.4--4.0)}\\
\textit{blnt\_shrp\_ln} (\textit{ang\_tri\_ln}$+$\textit{crs\_ncrs}) & 5.0 (4.3--5.1) & 4.4 (4.1--4.6) & \textbf{4.0 (3.9--4.3)}\\
\textit{ang\_crs} (\textit{ang\_tri\_ln}$+$\textit{blnt\_shrp\_ln}) & 3.2 (3.0--3.9) & \textbf{3.1 (2.8--3.3)} & \textbf{3.1 (2.6--3.6)}\\
\textit{ang\_crs} (\textit{ang\_tri\_ln}$+$\textit{ang\_crs\_ln}) & \textbf{2.2 (2.1--2.6)} & \textbf{2.2 (1.8--2.7)} & \textbf{2.3 (2.0--2.7)} \\
\end{tabular}
\end{table}

\subsubsection{Adding blocks to triplets of networks} The results of adding blocks to triplets of networks are presented in Table~4. They show that having three original networks improved the performance of the block network. For example, using three original networks (\textit{ang\_tri\_ln}$+$\textit{crs\_ncrs}$+$\textit{blnt\_shrp\_ln}) a block which is trained for \textit{ang\_crs} outperformed a similar network learned from scratch. Interestingly, for \textit{blnt\_shrp\_ln}, combining multiple original networks (two or three) and adding a block with an empty first layer was more efficient than adding a block with 100 neurons in the first hidden layer to a single original network (compare Tables~1--4). The respective number of trained weights was also smaller in the former case.

\begin{table}
\label{table4}
\centering
\caption{Adding blocks to triplets of original networks.}
\begin{tabular}{l|c}
condition & 0-50-50\\\hline
\textit{ang\_crs} (\textit{ang\_tri\_ln}$+$\textit{crs\_ncrs}$+$\textit{blnt\_shrp}) & 3.2 (3.1--3.5)\\
\textit{ang\_crs} (\textit{ang\_tri\_ln}$+$\textit{ang\_crs\_ln}$+$\textit{crs\_ncrs}) & \textbf{2.3 (2.1--2.7)} \\
\textit{ang\_crs} (\textit{ang\_tri\_ln}$+$\textit{crs\_ncrs}$+$\textit{blnt\_shrp\_ln}) & \textbf{2.9 (2.7--3.2)}\\
\textit{ang\_crs\_ln} (\textit{ang\_tri\_ln}$+$\textit{crs\_ncrs}$+$\textit{blnt\_shrp\_ln}) & \textbf{7.8 (7.3--8.2)}\\
\textit{ang\_crs\_ln} (\textit{ang\_tri\_ln}$+$\textit{crs\_ncrs}$+$\textit{blnt\_shrp}) & 8.4 (8.2--8.6)\\
\textit{blnt\_shrp} (\textit{ang\_crs}$+$\textit{ang\_tri\_ln}$+$\textit{crs\_ncrs}) & \textbf{0.7 (0.7--0.8)}\\
\textit{blnt\_shrp} (\textit{ang\_crs\_ln}$+$\textit{ang\_tri\_ln}$+$\textit{crs\_ncrs}) & \textbf{0.7 (0.7--0.8)}\\
\textit{blnt\_shrp\_ln} (\textit{ang\_crs\_ln}$+$\textit{ang\_tri\_ln}$+$\textit{crs\_ncrs}) & \textbf{4.0 (3.5--4.1)}\\
\textit{blnt\_shrp\_ln} (\textit{ang\_crs}$+$\textit{ang\_tri\_ln}$+$\textit{crs\_ncrs}) & \textbf{3.9 (3.4--4.2)}\\
\end{tabular}
\end{table}

\section{Conclusions}
Our results suggest that adding blocks to neural networks can be an efficient way to obtain networks capable of performing several tasks. In certain cases such composite networks outperform the networks trained from scratch, while having almost one order of magnitude fewer weights. Also, we observed a smaller range of variability in performance for block networks when compared to the original networks. Adding fewer new weights offers a significant gain in computational time if the stimuli are to be tested for all tasks. The performance of the block networks can be partly explained by the pre-training effect; the original network trained on its task can be considered as a pre-trained sub-network for a new task, to which the added block is trained. Another reason for better performance could be the increase in available training data. Since the tasks share some similarities, certain features could be learned from both original and new datasets.  Pilot computational experiments show that the improvement offered by the block networks becomes more significant when the amount of training data is reduced for both the original and new tasks. This ability to re-use features from different datasets may be beneficial for practical problems where it is often difficult to collect a large amount of training data for a specific task.

\subsubsection*{Acknowledgments.} This work was funded by the European Research Council (FP 7 Program) ERC Advanced Grant ``FEEL'' to KO'R

\bibliography{lm}{}
\bibliographystyle{splncs03}

\end{document}